\pdfoutput=1
\documentclass[11pt]{article}
\usepackage[preprint]{acl}
\usepackage{times}
\usepackage{latexsym}
\usepackage[T1]{fontenc}
\usepackage[utf8]{inputenc}
\usepackage{microtype}
\usepackage{inconsolata}
\usepackage{graphicx}
\usepackage{subcaption}
\usepackage{booktabs}
\usepackage{amsmath}
\usepackage{paralist}
\usepackage{multirow}

\title{Enhancing RLHF with Human Gaze Modeling}

\author{
 \textbf{Karim Galliamov,\textsuperscript{1}}
   \textbf{Ivan Titov,\textsuperscript{2}}
   and 
 \textbf{Ilya Pershin\textsuperscript{1}}
\\
 \textsuperscript{1}Innopolis University,
 \textsuperscript{2}University of Edinburgh
\\
 \small{
   \href{mailto:k.galliamov@innopolis.university}{k.galliamov@innopolis.university,}
 }
 \small{
 \href{mailto:ititov@inf.ed.ac.uk} {ititov@inf.ed.ac.uk,}
 }
 \small{
    \href{mailto:i.pershin@innopolis.ru}{i.pershin@innopolis.ru}
 }
}

\begin{document}
\maketitle
\begin{abstract}
Reinforcement Learning from Human Feedback (RLHF) aligns language models with human preferences but  is computationally expensive. We explore two approaches that leverage human gaze modeling to enhance RLHF: (1) gaze-aware reward models and (2) gaze-based distribution of sparse rewards at token level. Our experiments demonstate that gaze-informed RLHF achieves faster convergence while maintaining or slightly improving performance, thus, reducing computational costs during policy optimization. These results show that human gaze provides a valuable and underused signal for policy optimization, pointing to a promising direction for improving RLHF efficiency.
\end{abstract}

\section{Introduction}

Reinforcement Learning from Human Feedback (RLHF) has emerged as a powerful paradigm for aligning language models with human values and preferences \cite{ziegler2019fine}, typically involving training a reward model on human preference data,  followed by policy optimization algorithms such as PPO \cite{schulman2017proximal}.  
The process demands substantial computational resources and time, as it typically requires a large number of optimization steps to reach convergence.   This inefficiency is driven in part by  a fundamental challenge in RLHF: the sparsity of feedback signals \cite{chan2024dense}. Reward models typically provide a single scalar value for an entire sequence, offering limited guidance on which specific parts of a model's output contribute positively or negatively to human preferences.

Human visual attention -- as captured by eye tracking and gaze patterns -- 
offers fine-grained insights into where cognitive effort is directed during text processing \cite{human_gaze_review}. When skimming through text, humans naturally fixate on segments that are particularly informative, challenging, or salient.  
Incorporating gaze information into RLHF may help identify which parts of a sequence drive human judgments, thus improving the efficiency  policy training.

 We leverage a gaze prediction model trained on existing eye-tracking corpora and investigate two alternative approaches for integrating its predictions into the RLHF pipeline. To our knowledge, this is the first work to explore the use of human gaze modeling within the RLHF framework:

\begin{itemize}

\item First, we evaluate the application of gaze-informed reward models, introduced in \citet{lopez-cardona2025seeing}. These models have demonstrated improved accuracy in preference ranking -- i.e. ranking outputs according to human preferences -- but have not yet been evaluated within RLHF pipelines. Our contribution is to show that incorporating these reward models into PPO and GRPO frameworks \cite{shao2024deepseekmath} leads to faster convergence. We refer to this approach as \textit{GazeRM}.

\item Second, we propose \textit{GazeDistrib}. Inspired by prior work that leverages reward model attention for reward attribution \citep{chan2024dense}, our approach instead relies on predicted human gaze to assign token-level credit in RLHF.  By aligning rewards with human attention, we aim to provide more targeted feedback, accelerating convergence and potentially improving performance. Notably,  \textit{GazeDistrib}  requires no changes to the reward model itself -- only to how its scores are distributed.
\end{itemize}

Our experiments with both PPO and GRPO demonstrate that gaze-informed RLHF consistently achieves faster convergence—by a factor of 1.5-2 — compared to standard approaches. This acceleration occurs while maintaining or slightly improving final performance, suggesting that human visual attention patterns provide valuable signals for more efficient learning.

The primary contributions of this work are:

\begin{compactitem}
    \item integrating human gaze prediction models into RLHF pipelines through two alternative methods;
    \item empirical evidence showing significant convergence acceleration across two online-RL algorithms;
    \item analysis of gaze-informed reward distribution effects on the learning process.\footnote{The code is available at \url{https://github.com/KGallyamov/Gaze_RLHF}}
\end{compactitem}

\begin{figure}
  \begin{subfigure}[b]{\linewidth}
    \centering
    \includegraphics[width=0.98\linewidth,]{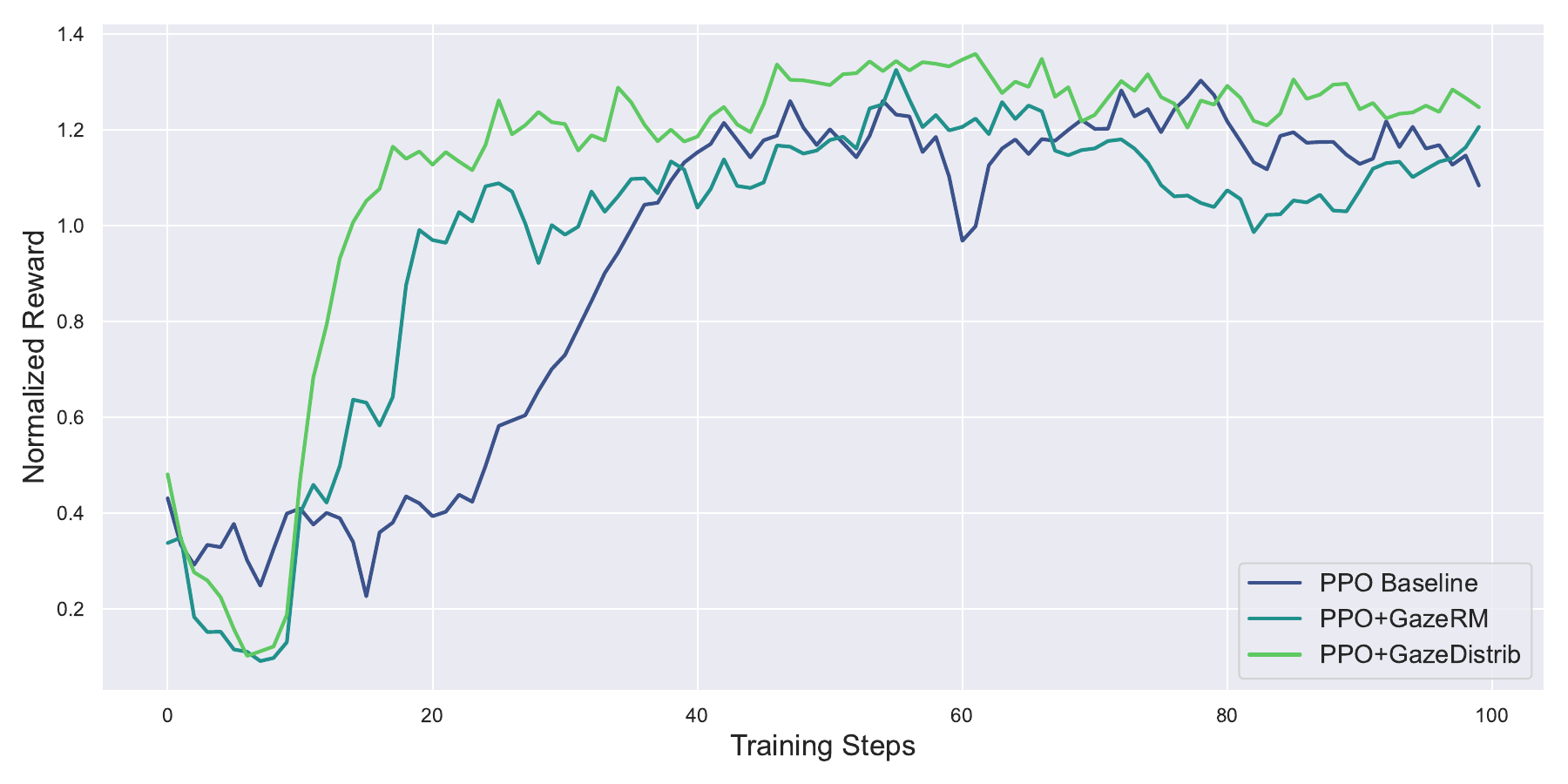}
    \caption{Validation scores gained by policy in PPO}
    \label{fig:ppo-holdout-rewards}
  \end{subfigure}


  \centering
  \begin{subfigure}[b]{\linewidth}
    \centering
    \includegraphics[width=0.98\linewidth]{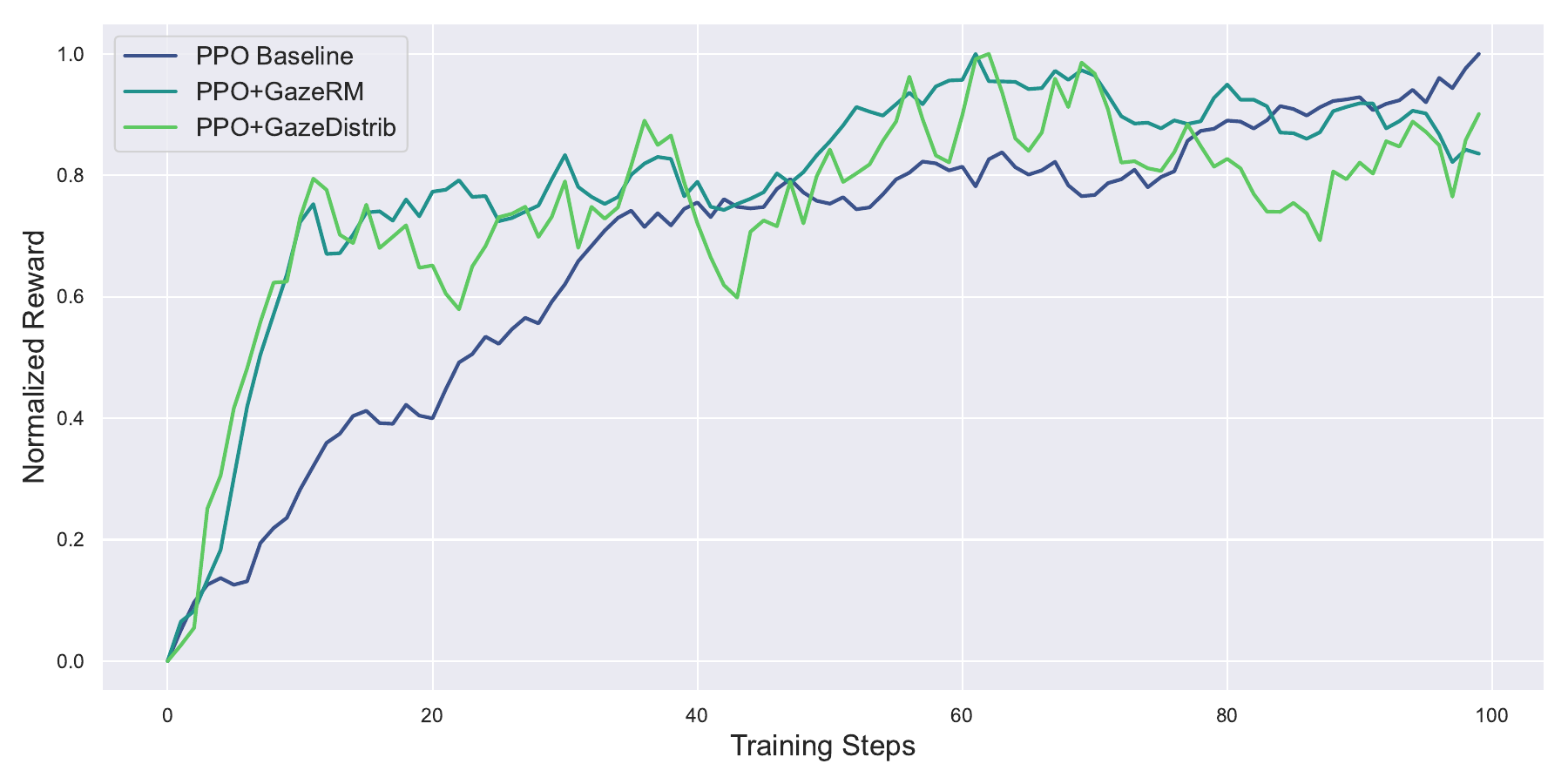}
    \caption{Training rewards gained by policy in PPO}
    \label{fig:ppo-training-rewards}
  \end{subfigure}

  \caption{Performance of LLaMa-7B-open-instruct models trained with and without gaze information on HH-RLHF}
  \vspace{-1.8em}
  \label{fig:ppo-rewards}
  
\end{figure}

\section{Background and Related Work}

\subsection{Gaze Prediction Models}
Collecting human gaze data is time-consuming and costly. A practical alternative is to train a model on existing eye-tracking corpora \citep{gaze_dataset_one, gaze_dataset_two} and use it to predict gaze patterns on new data. 
Such models predict various token-level features including First Fixation Duration (how long a reader initially fixates on a token), Go-Past Time (total time spent from first encountering a token until moving past it), Total Reading Time (cumulative duration of all fixations on a token), and number of fixations (nFix). The predictions serve as proxies for actual human visual attention patterns.

\subsection{Gaze-Informed Reward Models}
\citet{lopez-cardona2025seeing} proposed integrating gaze features into reward modeling by projecting them into the model's hidden representation space using a trainable feedforward neural network. These gaze feature projections are either added to the first layer's hidden activations or concatenated. 

Their experiments demonstrated substantial improvements in preference prediction accuracy with both integration methods. However, their work did not extend to evaluating how these gaze-enhanced reward models perform when incorporated into complete RLHF training pipelines.


\subsection{Dense Reward Distribution from Model Attention}
Previous work by \citet{chan2024dense} explored an alternative to sparse sequence-level rewards by proposing a method to distribute rewards at the token level using the attention patterns from a reward model's final layer. Their approach transforms the single scalar reward into a dense, token-specific signal that provides more granular feedback during training. Their experiments demonstrated that such dense reward distribution leads to faster convergence compared to traditional sparse reward approaches, suggesting that more fine-grained feedback signals can improve the efficiency of policy optimization. However, this method  has no explicit connection to human attention.

\section{Our approach}
\label{sec:our_approach}

\subsection{Human attention model analysis}

To get an intuition for why modeling human gaze may improve training dynamics, we analyze how gaze is distributed across different parts of speech (PoS).  We applied PoS tagging using FLAIR \citep{akbik2018coling} to obtain grammatical categories for each word. Since our gaze prediction model operates on subword tokens while PoS tags are assigned at the word level, we computed word-level attention scores by summing the predicted attention values across all subtokens comprising each word.

In Table \ref{tab:gaze_analysis}, we report the mean human attention scores by part of speech (PoS), using coarser categories derived from the Penn Treebank tags used by FLAIR. We computed word-level human attention scores for each instruction in OASST2, applied PoS tagging, and aggregated the scores by grammatical category to obtain the mean attention value for each PoS across all word instances.

Overall, the human gaze model clearly attends to content-rich  words (especially nouns and verbs), as well as punctuation. One possible explanation for \textit{GazeDistrib}’s faster convergence is its strategy of assigning higher rewards to tokens that carry more information. By prioritizing these key elements, the model can more efficiently capture underlying patterns and structures, leading to faster and more effective training.

\begin{table}[ht]
\centering
\small
\footnotesize 
\begin{tabular}{|l|l|c|}
\hline
\textbf{POS Tags} & \textbf{Meaning} & \textbf{Gaze} \\
\hline
TO & to & 0.0122 \\
NFP & Superfluous punctuation & 0.0246 \\
AFX & Affix & 0.0282 \\
SYM & Symbol & 0.0291 \\
CC & Coordinating conjunction & 0.0318 \\
EX & Existential there & 0.0331 \\
XX & Unknown & 0.0369 \\
DT & Determiner & 0.0376 \\
IN & Preposition or conjunction & 0.0386 \\
FW & Foreign word & 0.0397 \\
PRP & Personal pronoun & 0.0402 \\
LS & List item marker & 0.0419 \\
MD & Modal & 0.0422 \\
WDT & Wh-determiner & 0.0437 \\
CD & Cardinal number & 0.0442 \\
UH & Interjection & 0.0578 \\
WRB & Wh-adverb & 0.0620 \\
WP & Wh-pronoun & 0.0748 \\
WP\$, PRP\$ & Possessive (wh-)pronouns & 0.0854 \\
. | , | HYPH & Punctuation marks & 0.1316 \\
JJ* & Adjectives & 0.1355 \\
RB* & Adverbs & 0.1466 \\
NN* & Nouns & 0.2295 \\
VB* & Verbs and gerunds & 0.2697 \\
\hline
\end{tabular}
\caption{Part-of-Speech Tags Gaze Scores}
\vspace{-2em}
\label{tab:gaze_analysis}
\end{table}

We investigate two alternative approaches to integrating human gaze information into RLHF pipelines: enhancing reward models with explicit gaze features and using human gaze patterns to redistribute sparse sequence-level rewards across individual tokens.

\begin{figure}
  \begin{subfigure}[b]{\linewidth}
    \centering
    \includegraphics[width=0.98\linewidth,]{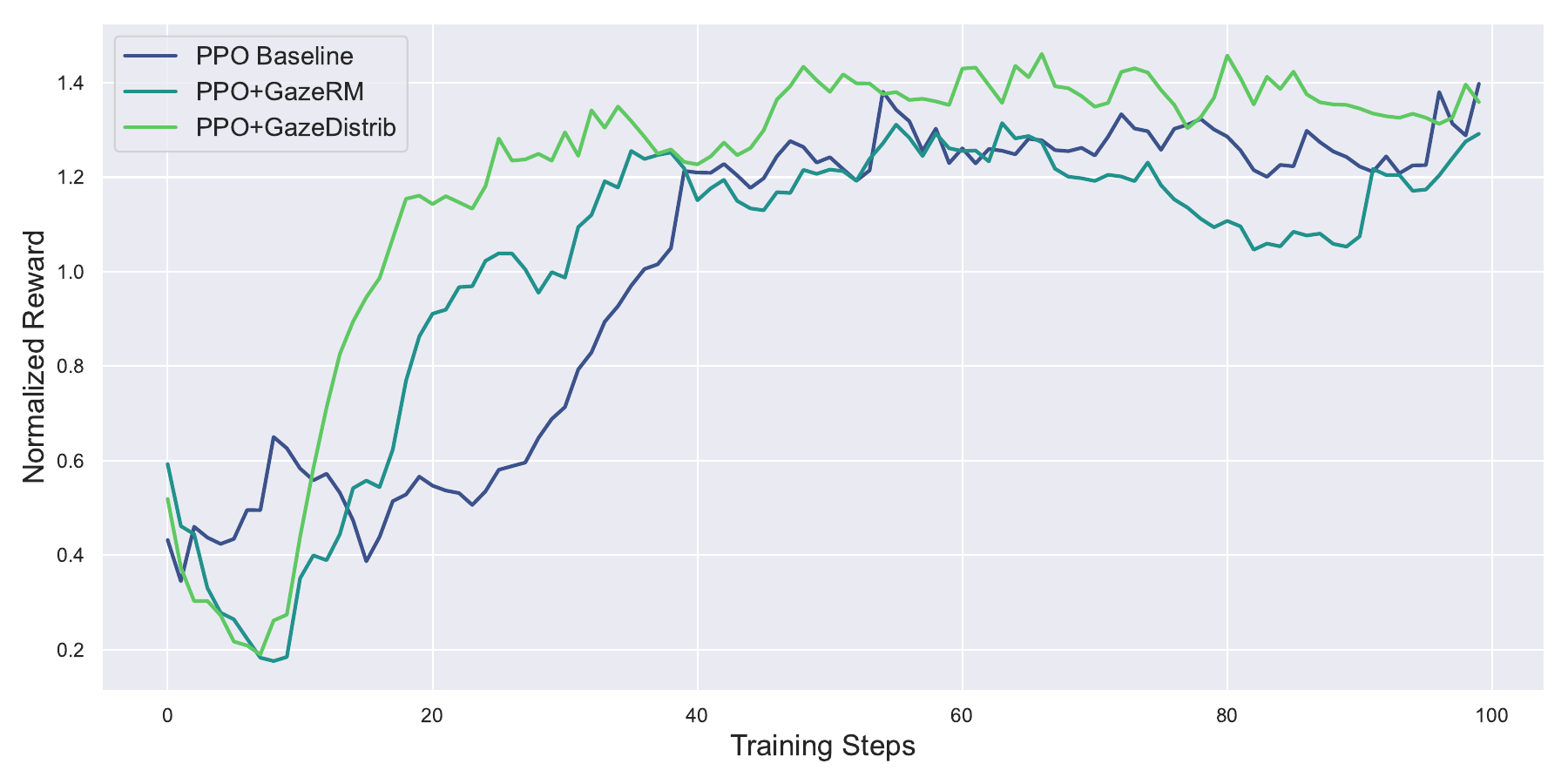}
    \caption{Validation scores gained by policy in PPO}
    \label{fig:ppo-holdout-rewards-diff-sft}
  \end{subfigure}


  \centering
  \begin{subfigure}[b]{\linewidth}
    \centering
    \includegraphics[width=0.98\linewidth]{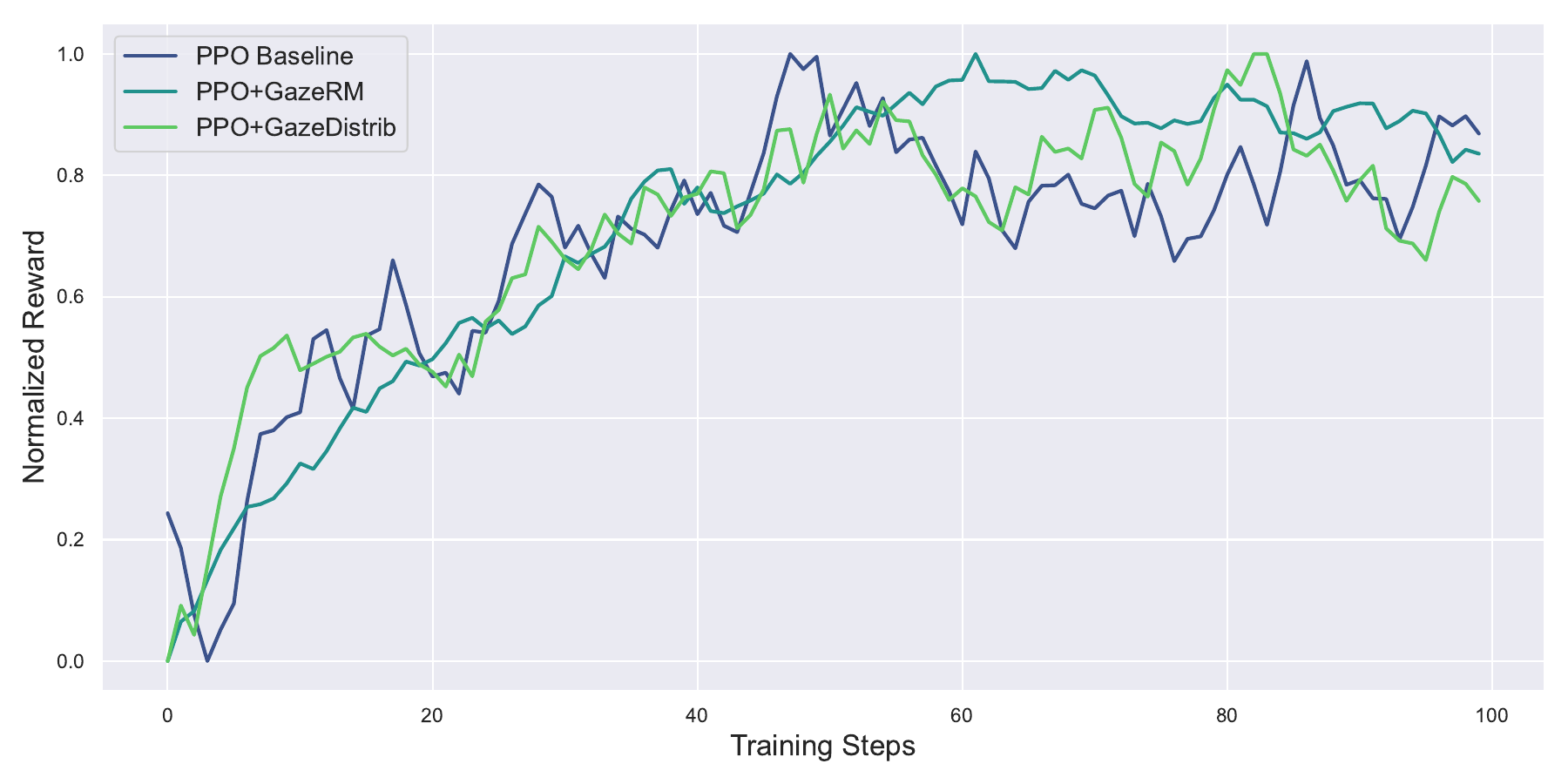}
    \caption{Training rewards gained by policy in PPO}
    \label{fig:ppo-training-rewards-diff-sft}
  \end{subfigure}

  \caption{Performance of Mistral-7B trained with and without gaze information }
  \vspace{-1.8em}
  \label{fig:ppo-rewards-diff-sft}
  
\end{figure}

\subsection{Gaze-Informed Reward Models}

Building upon previous work by \citet{lopez-cardona2025seeing}, 
the human eye tracking data is accomplished by projecting the predicted gaze features into the hidden space of a reward model. These projections are either added or concatenated directly into the first layer token embeddings of the reward model.

\citet{lopez-cardona2025seeing} only evaluated their models on reward benchmarks and validation preference datasets. While they observed improved performance compared to gaze-agnostic baseline reward models, they did not integrate their reward models into the  RLHF pipeline. Our work examines whether these improvements translate into more efficient and effective policy optimization when deployed within the complete RLHF framework. Our central hypothesis is that a more precise and informative reward signal derived from human attention patterns will lead to more targeted and efficient updates during the policy optimization phase, thereby accelerating convergence and potentially improving the quality of the final policy. Samples of predicted human gaze could be found in Appendix \ref{sec:appendix_human_gaze}.

\subsection{Dense reward from human gaze}

Traditional RLHF methods commonly assign a single scalar reward to an entire generated sequence, providing a very sparse signal that offers limited guidance on which specific tokens or spans contributed positively or negatively to the overall human judgement. We address this sparsity problem by leveraging human gaze predictions to infer token-level importance and distribute rewards accordingly.

In our implementation, we use Total Reading Time (TRT) as the primary gaze metric for distributing rewards at the token level. TRT measures the cumulative duration a reader fixates on each token,  including both initial reading and any subsequent rereadings. This makes it a strong proxy for a token’s processing load and its importance in human evaluation. By levereging TRT, we can derive token-level dense rewards using readily available gaze prediction models, offering a more fine-grained signal for training.

Given a predicted TRT distribution $T = \{t_1, t_2, ..., t_n\}$ for a sequence of $n$ tokens and a scalar reward $R$ for the entire sequence (obtained from standard human preference judgments), we compute token-level rewards $r_i$ using a softmax-like distribution over the gaze values:

\begin{equation}
r_i = R \cdot \frac{e^{t_i}}{\sum_{j=1}^{n} e^{t_j}}
\end{equation}

This approach (\textit{GazeDistrib}) ensures that tokens receiving more visual attention (higher TRT) from humans are assigned proportionally larger portions of the overall sequence reward $R$. This process creates a dense reward signal that is aligned with human attention and cognitive effort, providing more precise and granular guidance to the model optimization process compared to a single sequence-level scalar.

Crucially, since the used gaze predicting models are significantly smaller (up to 125M parameters) compared to the backbone models used for both the reward model and the trained policy (7B parameters), their inference cost and storage requirements are negligible and do not represent a bottleneck in either of the proposed integration scenarios.

\section{Experiments}

We train our models on HH-RLHF \citep{2022arXiv220405862B}. This is a dataset provided by Anthropic, totaling 161K human annotated chosen-rejected dialog pairs. We also run our experiments on OASST-2 \citep{kopf2023openassistant} - another preference pairs dataset, with 64K prompts. 
For RLHF alignment, only instruction prompts are required — not full dialogues — so we truncate each example before the model's final response. 
Following \citet{lopez-cardona2025seeing}, we filter out all non-English examples from the dataset, as the gaze predicting model - and thus the gaze-informed reward model - was only trained on English corpora. To demonstrate the robustness of our approach, we integrate both our methods with both PPO and GRPO methods. The implementation of these methods is taken from TRL \cite{vonwerra2022trl}.

\subsection{Gaze Modeling}

In our work, we used gaze models from \citet{gaze_dataset_one} and \citet{ gaze_dataset_two}. The models  are trained on large corpora with gaze annotations; the token-level regression tasks involved predicting several variables such as First Fixation Duration, Go-Past Time, and Total Reading Time. After training, the models are frozen. During our experiments, we use these models to infer gaze features, which are then used as scores for dense reward on policy model's output generations. Other way of gaze features utilization is to train the reward models after having these gaze features computed for the preference corpora as in \citep{lopez-cardona2025seeing}. 

\subsection{Reward models}

We reproduce the work of \citet{lopez-cardona2025seeing} by training Mistral 7B \citep{jiang2024identifying} for reward modeling with concatenation of gaze projected features to the input token embeddings (setup referred to as $f_{comb_{2.5}}$ in the original paper).

\subsection{Evaluation}

Since our experiments involve policies trained with different reward models, directly comparing their final reward values would be misleading. To enable meaningful cross-training comparisons, we utilized a ``hold-out'' reward model - the one not used in any policy training process - to evaluate all policies on a consistent benchmark. We use OpenAssistant's deberta-large-v2 \citep{kopf2023openassistant} reward model as this validating model. The selected reward model achieves accuracy of $69.25$ on HH-RLHF validation split thus making it quite a strong model for comparing models on that dataset.
This hold-out model serves two purposes: it provides comparable scores across different training runs and validates that any observed improvements in convergence speed reflect genuine learning rather than simply exploiting weaknesses in specific reward schemes, i.e. reward hacking. We define a policy's validation score as the difference between its hold-out reward score and the score of the initial SFT model. For further analysis, we also plot min-max normalized plots. 

All plots and tables show averages over 3 runs with different seeds to mitigate the stochasticity in policy optimization.

\section{Results}

\begin{table*}[ht]
    \centering
    \small
    \footnotesize 
    \begin{tabular}{@{}lcccc@{}}
        \toprule
        \textbf{Method} & \multicolumn{2}{c}{\textbf{LLaMa (HH-RLHF)}} & \multicolumn{2}{c}{\textbf{Mistral (OASST2)}} \\
        \cmidrule(lr){2-3} \cmidrule(lr){4-5}
        & \textbf{Val. Score} & \textbf{Steps to Conv.} & \textbf{Val. Score} & \textbf{Steps to Conv.} \\
        \midrule
        Baseline & $1.09 \pm 0.08$ & $41.40 \pm 0.21$ & $1.21 \pm 0.15$ & $53.40 \pm 0.29$ \\
        GazeRM & $1.19 \pm 0.10$ & $28.33 \pm 0.54$ & $1.10 \pm 0.31$ & $29.20 \pm 0.95$ \\
        GazeDistrib & $1.22 \pm 0.12$ & $22.33 \pm 1.02$ & $1.16 \pm 0.25$ & $30.60 \pm 0.73$ \\
        \bottomrule
    \end{tabular}
    \caption{Convergence and quality comparison on PPO across different model architectures and datasets}
    \label{tab:approaches_comparison}
    \vspace{-1em}
\end{table*}

Figure \ref{fig:ppo-rewards} shows the convergence behavior of PPO with and without gaze-based rewards across both model architectures. The results consistently demonstrate significant speed-up in convergence: validation scores grow much faster for gaze-aware approaches in both LLaMa and Mistral experiments. However, they ultimately reach similar performance levels, suggesting that gaze-aware methods primarily accelerate training rather than improve final model quality.

The training reward behavior varies across different SFT initializations. For one policy initialization, training rewards closely resemble validation scores \ref{fig:ppo-training-rewards}, while for another initialization method \ref{fig:ppo-training-rewards-diff-sft}, there is no significant difference in training rewards between methods. This inconsistency in min-max normalized reward model comparisons confirms that tracking validation scores provides more reliable insights.

To demonstrate that our approaches are invariant to policy optimization algorithms, we repeated the PPO experiments using GRPO \citep{shao2024deepseekmath}, observing similar convergence speed improvements of approximately 1.3 times faster with gaze-guided methods. The consistency across different model architectures (LLaMa and Mistral) and datasets (HH-RLHF and OASST2) validates the generalizability of our approach. When using Mistral, we aligned the reward model backbone with the corresponding architecture to ensure tokenizer compatibility, which is crucial for dense reward training. For additional details on GRPO experiments, refer to Appendix \ref{sec:appendix_grpo}.

\section{Conclusion}
In this paper, we have presented a novel method of integrating human gaze data into RLHF through dense reward attribution guided by human attention patterns (GazeDistrib). Additionally, we validated that existing gaze-informed reward models from prior work, when incorporated into PPO and GRPO frameworks, provide substantial benefits to the training process. Our experiments demonstrate that both approaches accelerate convergence during policy optimization by a factor of 1.5-2 times without compromising final model performance.

The results indicate that human visual attention patterns provide valuable signals that enhance alignment efficiency. By addressing the sparse reward problem through more targeted feedback, our gaze-based approaches offer a promising direction for improving computational efficiency in RLHF.

\section*{Limitations}

In this work, we have experimented only with on-policy RLHF methods. Further experiments with off-policy methods, such as DPO, are required.  Our study was limited to two datasets initialization and performed only on English. Future work should extend these experiments to additional language, datasets, and language models. 

\bibliography{custom}

\appendix

\section{Implementation details and hparams tuning}
\label{sec:appendix_hyperparameters}

In this section, we list the setups used for PPO and GRPO experiments: SFT initialization checkpoints, hold-out reward models and datasets.



\subsection{}

\begin{description}
    \item[Dataset:] \href{https://huggingface.co/datasets/Anthropic/hh-rlhf}{Anthropic/hh-rlhf}
    \item[Holdout Reward Model:] \href{https://huggingface.co/OpenAssistant/reward-model-deberta-v3-large-v2}{OpenAssistant/reward-model-deberta-v3-large-v2}
    \item[Policy Model Initialization:] 
    \href{https://huggingface.co/VMware/open-llama-7b-open-instruct}{VMware/open-llama-7b-open-instruct}
\end{description} 

\subsection{}

\begin{description}
    \item[Dataset:] \href{https://huggingface.co/datasets/OpenAssistant/oasst2}{OASST2}
    \item[Holdout Reward Model:] \href{https://huggingface.co/OpenAssistant/reward-model-deberta-v3-large-v2}{OpenAssistant/reward-model-deberta-v3-large-v2}
    \item[Policy Model Initialization:] 
    \href{https://huggingface.co/mistralai/Mistral-7B-v0.1}{Mistral-7B-v0.1}
\end{description}

We run a sweep for learning rate and batch size identical to that described in \citep{chan2024dense}. 

\section{Transferring to GRPO}
\label{sec:appendix_grpo}

\begin{figure}
  \centering
  \begin{subfigure}[b]{\linewidth}
    \centering
    \includegraphics[width=0.98\linewidth]{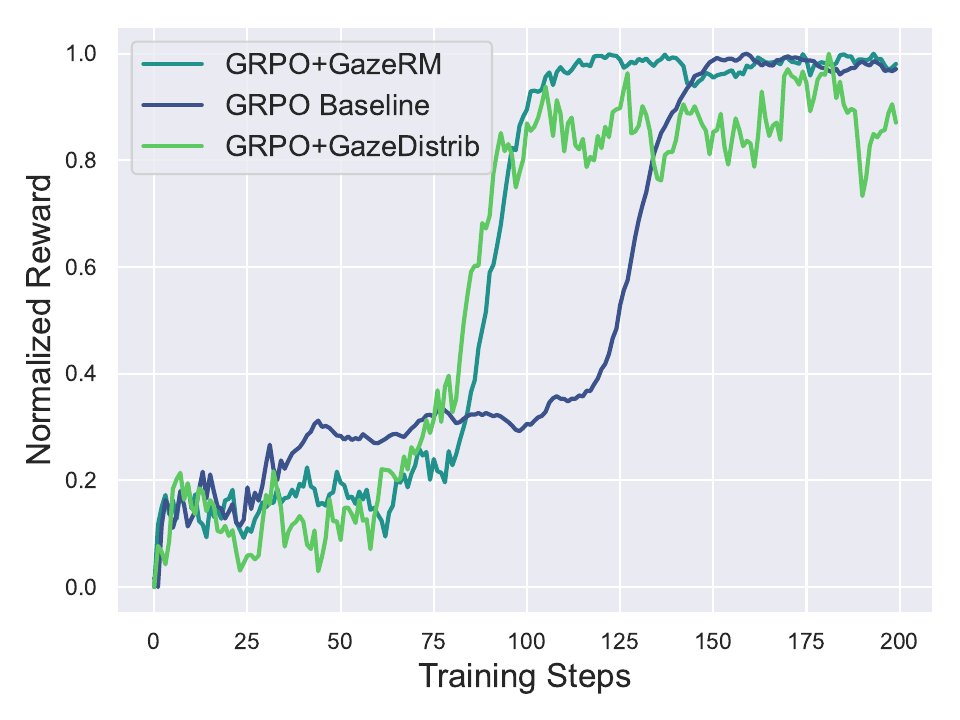}
    \caption{Training rewards gained by policy in GRPO}
    \label{fig:grpo-training-rewards}
  \end{subfigure}
  
  \vspace{-1.5em}
\end{figure}

On \ref{fig:grpo-training-rewards} one may see that the convergence speed boost observed on PPO also transfers to GRPO - however, the gained reward graph differs from that observed in PPO

\section{Sample human gaze patterns}
\label{sec:appendix_human_gaze}

\begin{figure*}[t]
  \includegraphics[width=0.96\linewidth]{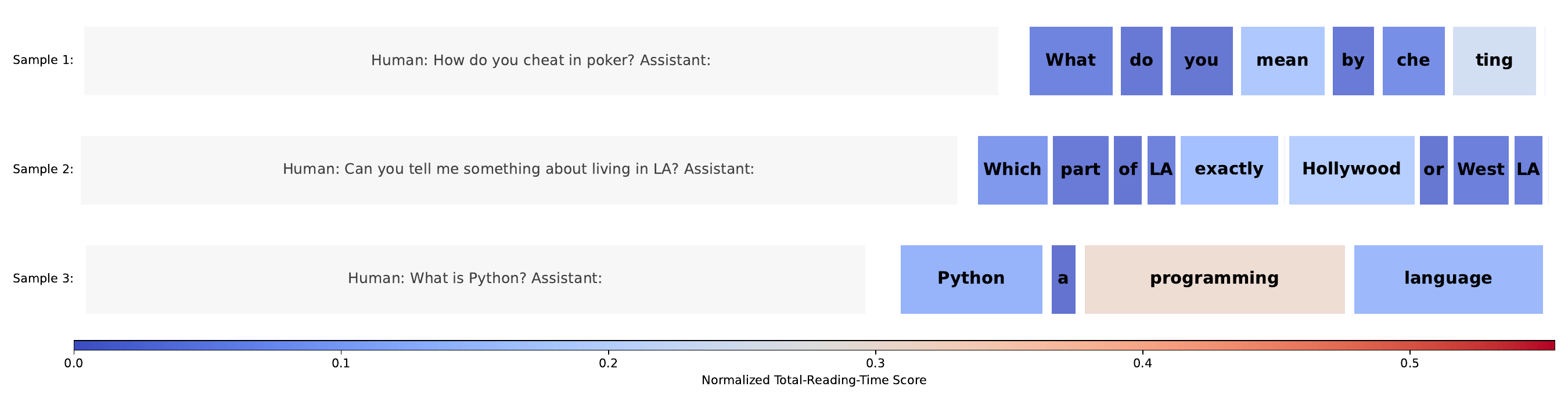}
  \caption {Samples of human gaze modeling}
  \vspace{-1.5em}
\label{fig:gaze-distribution-samples}
\end{figure*}

\end{document}